\documentclass[11pt]{article}

\usepackage[utf8]{inputenc}
\usepackage[T1]{fontenc}
\usepackage{lmodern}
\usepackage{geometry}
\usepackage{amsmath,amsfonts,amssymb,amsthm,mathtools}

\usepackage{graphicx}
\usepackage{subcaption}
\usepackage{caption}

\usepackage{algorithm}
\usepackage{algpseudocode}

\usepackage{booktabs}
\usepackage{multirow}

\usepackage{tikz-cd}

\usepackage{xcolor}
\usepackage{listings}

\usepackage{comment}

\definecolor{codegreen}{rgb}{0,0.6,0}
\definecolor{codegray}{rgb}{0.5,0.5,0.5}
\definecolor{codepurple}{rgb}{0.58,0,0.82}
\definecolor{backcolour}{rgb}{0.95,0.95,0.92}

\lstdefinestyle{mystyle}{
    backgroundcolor=\color{backcolour},
    commentstyle=\color{codegreen},
    keywordstyle=\color{magenta},
    numberstyle=\tiny\color{codegray},
    stringstyle=\color{codepurple},
    basicstyle=\ttfamily\footnotesize,
    breaklines=true,
    numbers=left,
    numbersep=5pt,
    showstringspaces=false,
    tabsize=2
}
\usepackage{hyperref}

\usepackage{authblk}

\newtheorem{theorem}{Theorem}
\newtheorem{definition}{Definition}
\newtheorem{remark}{Remark}

\newcommand{\ma}{\textcolor{black}}
\newcommand{\bl}{\textcolor{black}}

\title{Density Matrix-based Dynamics\\
for Quantum Robotic Swarms}

\author[1,2,3]{Maria Mannone\thanks{Corresponding author: maria.mannone@icar.cnr.it}}
\author[4]{Mahathi Anand}
\author[5,6]{Peppino Fazio}
\author[7]{Abdalla Swikir}

\affil[1]{ICAR, National Research Council (CNR), Palermo, Italy}
\affil[2]{Institute of Physics and Astronomy, University of Potsdam, Germany}
\affil[3] {Potsdam Institute for Climate Impact Research (PIK), Member of the Leibniz Association, Germany}
\affil[4]{Learning Systems and Robotics Lab, Technical University of Munich (TUM), Germany}
\affil[5]{Department of Molecular Sciences and Nanosystems (DSMN), Ca' Foscari University of Venice, Italy}
\affil[6]{VSB, Technical University of Ostrava, Czechia}
\affil[7]{Mohamed bin Zayed University of Artificial Intelligence (MBZUAI), Abu Dhabi, UAE}

\date{} 

\begin{document}
 
\maketitle

\begin{abstract}
In a robotic swarm, parameters such as position and proximity to the target can be described in terms of probability amplitudes. This idea led to recent studies on a quantum approach to the definition of the swarm, including a block-matrix representation. \ma{However, the size of such matrix-based representation increases drastically with the swarm size, making them impractical for large swarms. Hence, in this work, we propose a new approach for modeling robotic swarms and robotic networks by considering them as mixed quantum states that can be represented mathematically via density matrices. The size of such an approach only depends on the available degrees of freedom of the robot, and not its swarm size and thus scales well to large swarms. Moreover, it also enables the extraction of local information of the robots from the global swarm information contained in the density matrices, facilitating decentralized behavior that aligns with the collective swarm behavior. Our approach is validated on several simulations including large-scale swarms of up to $1000$ robots. Finally, we provide some directions for future research that could potentially widen the impact of our approach.}
\end{abstract}



\section{Introduction}



\ma{\textcolor{black}{At the frontier of classical and quantum physics: defining a swarm of robots small enough to operate under quantum effects is a challenging task.
Here, the considered  problem is to obtain a suitable description of the swarm as a whole system composed of individual robots. As such, the domain of swarm robotics has} piqued considerable interest over the last decade.}
Much of this attention is attributed to the deployment of multi-agent robot teams to accomplish a
wide variety of \bl{challenging} tasks such as target monitoring, surveillance, environmental sensing, disaster recovery and exploration \cite{tlc_survey1}. \ma{On the other hand, the development of several bio-inspired algorithms~\cite{bio-main,bio_1,bio_2,bio_3} including deployment of robot swarms inspired by a flock of birds~\cite{oung}, insects~\cite{schranz,swarm_robotics}, fish~\cite{aquatic_swarm}, and even the microscopic world~\cite{sitti} has led to the progressive miniaturization of robotic swarms, making it possible to advance robotic solutions in a wide range of applications including medical technology~\cite{med_tech1,med_tech_2} and telecommunications~\cite{wireless01, wireless02}.}
However, the smaller the robotic swarm, the harder it is to control. \ma{In particular, as systems approach the nanoscale, quantum effects start to dominate, interactions between agents and their environment are influenced by quantum phenomena, and systems can no longer be explained by classical physics. As a result, controlling such robotic swarms is a challenging problem. Moreover, standard control algorithms require precise physical modeling and can no longer be applied to robotic swarms that operate under quantum effects. Therefore, it is critical to account for quantum-mechanical effects when modeling and reasoning about nano robotic swarms~\cite{quantum_nano}.}

\ma{The field of quantum computing uses quantum bits (\textit{qubit}) to extend the values of the standard $0,\,1$ to all the real values comprised between them (the interval $[0,1])$ \cite{stolze}. The paradigm offers a significant computational advantage to the traditional classical computing by taking advantage of quantum-mechanical principles including superposition and entanglement, and has been utilized for modeling and simulating forward kinematics of industrial robot manipulators~\cite{fazilat_novel_2022, otani_quantum_2025}, human robot interaction~\cite{quantum_human_robot, quantum_human_thinking}, and robot perception~\cite{Lanza_2020,Lanza_2021}. Moreover, quantum computing has been central in the advancement of machine learning techniques~\cite{wichert,QRL1,QRL2}, motion planning~\cite{quantum_mp, quantum_rrt, quantum_planning}, and optimization~\cite{qian_optimal_2019, otani_quantum_2025}, and has also been applied in the context of swarm robotics~\cite{koukam,quantum_fish,qRobot}.}
\bl{However there are significant challenges in utilizing quantum algorithms in the context of swarm robotics. For example, problems concerning stability~\cite{dasgupta_stability_2024} and robustness~\cite{berberich_robustness_2024} of the algorithms have not been well-investigated, though it has been suggested that algorithms such as quantum particle swarm optimization can lead to improvements in path-planning convergence~\cite{qian_optimal_2019}. This is due to the lack of appropriate modeling formalisms for quantum swarms. Therefore, in this article, we prepare the necessary ground for advanced control problems for quantum swarms by suggesting appropriate techniques to model them mathematically. }

\ma{Concerning robotic swarm modeling}, in \cite{swarm_paper}, a description of a robotic swarm is proposed in terms of a nested matrix whose diagonal submatrices contain the parameters of a single robot, while the off-diagonal elements contain the information exchange between the robots. Thus, the size of the matrix depends upon the degrees of freedom of each robot, the exchanged information, and the number of robots. In particular, the size of the matrix increases with the number of robots, which constitutes a computational problem for large swarms. \bl{ Though the article does not specifically concern quantum robots, the authors leverage quantum computing to quantize some information (e.g., position as well as proximity to the target) to model local interactions between the robots. Moreover, ~\cite{swarm_paper} only considers only the local-to-global approach, i.e., the global swarm behavior emerged from the local pairwise interactions, but not the vice-versa. }


\textcolor{black}{As a solution to the problems emerging in \cite{swarm_paper}, in this article we propose a different approach for modeling quantum robotic swarms. \ma{Particularly, we use the notion of \textit{density matrices}~\cite{density_matrix}, first introduced for the treatement of epistemic uncertainty on systems and their noisiness}, for describing robotic swarms. Particularly, we model a quantum robotic swarm as a mixed state given by a superposition of pure states that represent individual robots, mathematically indicating them with a density matrix. By shaping the density matrix for a robotic swarm, one can develop a suitable overall dynamics for the whole swarm as an operator acting on the corresponding density matrix.  It is important to note that in our approach, the size of the density matrices is rendered independent from the swarm size. In fact, it only depends on the number of degrees of freedom of the system, at the condition of treating robots as particles within a ``cloud'': the swarm is thus investigated as a whole, rather than as a collection of distinguishable objects. Moreover, information of local robots can also be extracted from the density matrices through partial trace operation, thus also facilitating a global-to-local approach in the modeling of quantum swarms.}

\bl{Another crucial difference of our approach with \cite{swarm_paper} is the neglection of interaction terms. In particular, while~\cite{swarm_paper} considers the pair-wise interaction terms between the robots, which contribute to the size of matrices obtained for the swarm, in our paper, we can neglect these interactions. This is rooted in the fact that the density matrices already consider the effect of the interaction terms through the mixing of pure quantum states. Therefore, our approach offers a scalable yet efficient solution for modeling the information contained in the swarm without introducing any unnecessary redundancies.}


\begin{figure}[ht!]
\centering
\includegraphics[width=0.8\columnwidth]{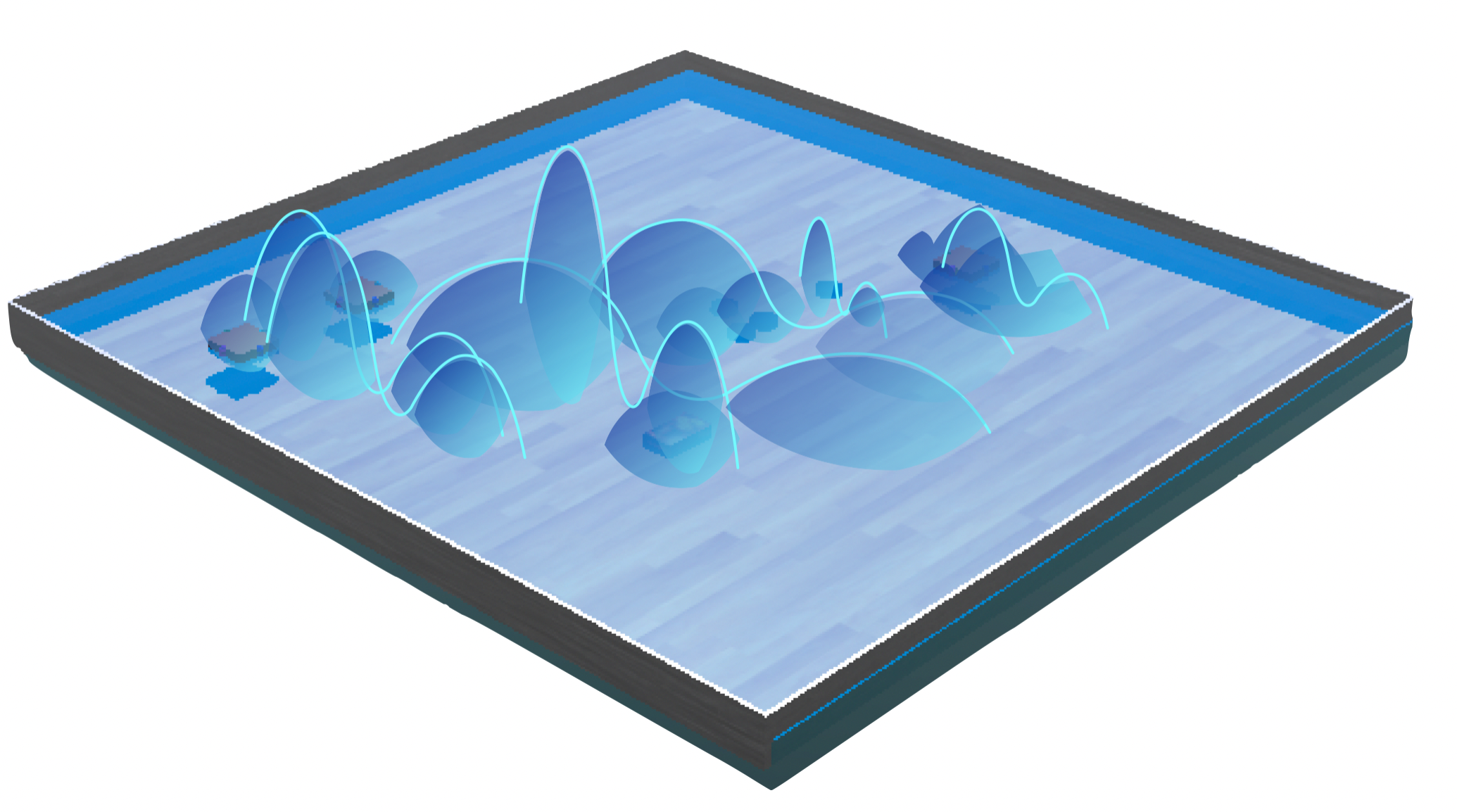}
\caption{Pictorial representation of the transition from a classic robotic swarm to a quantum-robotic swarm: in a given arena, each robot is considered in terms of a probability amplitude to be found in a certain space position.}
\label{robotic_waves}
\end{figure}



The article is organized as follows.
In Section \ref{bases} we present the necessary preliminaries from quantum computing, such as density matrices and their operations, that are useful for our results. 
\textcolor{black}{After having introduced the adopted quantum notation for single robots (Section \ref{single_robot_model}), }
 in Section \ref{new_model} we propose our modeling framework and present its advantages.
In Section \ref{examples} we demonstrate the application of our results through simple examples, \textcolor{black}{also showing their implementation in Python with the library Qiskit. In Section \ref{time_evolution} we discuss in detail the problem of time evolution, proposing a pseudocode of local-global swarm behavior, with a special reference to the network modeling in telecommunications, ending the section with a toy example of implementation.} Finally, in Section \ref{future} we summarize our results and present some extensions and applications of our approach to be considered in the future.

\section{Notations and Preliminaries}\label{bases}

\subsection{Notations}

\bl{The set of real numbers are denoted by $\mathbb{R}$, and $\mathbb{R}^n$ denotes the $n$-dimensional real space. The set of complex numbers is denoted by $\mathbb{C}$, and a complex matrix $A \in \mathbb{C}^{m \times n}$ is of size $m \times n$.
We use the standard bra-ket notation to describe a quantum state: $|\psi\rangle$ is the ket vector, a quantum pure state in a (complex) inner product vector space $V$ corresponding to an eigenstate of an operator, and $\langle\psi|$ is the bra vector, the complex conjugate of $|\psi\rangle$. Moreover, the notation $\langle\psi|\psi\rangle$ indicates the inner product, i.e., the product of bra and ket vectors, while $|\psi\rangle\langle\psi|$ denotes the outer product.}

\bl{For a matrix $A$, $Tr(A)$ is the trace operation, the sum of diagonal elements of a matrix. The operator $Tr$ indicates the Trace operation, defined as the sum of the diagonal elements of the matrix, that is $\sum_i T_{i,i}$, which is indicated as $Tr(T):=\Sigma_i\langle i|T|i\rangle$ in quantum mechanics. $A_t$ is a generic time-dependent operator; its complex conjugate is $A_t^{\dag}$, such that $(A^{\dag}_t)^{\dag}=A_t$.
If $A$ is Hermitian (self-adjoint), thus it is equal to its complex conjugate: $A_t^{\dag}=A_t$. Finally, we use $\otimes$ to describe the tensor product.} 


\subsection{Preliminaries} \label{sec:prelim}
In quantum mechanics, the behavior of a quantum state is described by the Schr\"{o}dinger equation~\cite{Schr01}:
\begin{equation}
i\hbar\frac{d}{dt}|\psi(t)\rangle=\mathbf{H}|\psi(t)\rangle,
\end{equation}
where $\mathbf{H}$ is the Hamiltonian operator,
$i$ is the imaginary unit, $\hbar$ is the \bl{reduced Planck constant} \textcolor{black}{that will be set equal to 1 here, as often done in quantum physics,} and $|\psi(t)\rangle$ represents the wavefunction \bl{at time t} as a mathematical description of the \emph{pure quantum state} in terms of its probability amplitude. 
\ma{In fact, a pure quantum state describes the position of a robot along its dimensional axes. For example, in case of one-dimensional movement, the $x$- position of a robot is given by}
\begin{equation}
\ma{|\psi(t)\rangle = \alpha^x |0\rangle + \beta^x |1 \rangle,}
\end{equation} \label{eq:purestate}
\ma{where $\alpha^x$ and $\beta^x$ are probability amplitudes along $|0\rangle$ (left) and $|1 \rangle$ (right), respectively with ${\alpha^x}^2 
+{\beta^x}^2 = 1$. 
Similarly, in the 2-dimensional (x-y) plane, one obtains the robot position as a pure state that is a superposition of several qubits describing the position along the $x$ and $y$ directions, i.e., as a superposition of the basis states $|00\rangle, |01\rangle, |10\rangle,$ and $|11\rangle$, similar to that defined in~\eqref{eq:purestate} (c.f. also~\eqref{caso_completo}).
} 
\bl{When dealing with an ensemble of quantum states, such as in a quantum robotics swarm \ma{consisting of multiple robots}, one considers a  mixture of different pure quantum states, known as \emph{mixed states}. In this case, one uses the more general master equation}~\cite{Naka01, nielsen}
\begin{equation}
\frac{d}{dt}\rho(t) = \mathcal{L}(\rho(t)),
\end{equation}
where $\rho(t)$ is the density operator, generalizing the mathematical description of \bl{the mixed quantum states, and $\mathcal{L}$ is the Lindblad operator given by 
\begin{equation} \label{lindblad_operator}
\begin{split}
&\mathcal{L}
(\rho(t))=-i[H,\rho(t)] + \\&\sum_{\alpha}\gamma_{\alpha}(t)\left(L^{T}\rho(t)L^{\ast}-\frac{1}{2}L^{\dag}L\rho(t)-\frac{1}{2}\rho(t)L^{\dag}L\right),
\end{split}
\end{equation}
where the operator $L$ represents the interaction \bl{of the system} with the environment.} Eq. \eqref{lindblad_operator} corresponds to a Markovian dynamics when $\gamma_\alpha \geq 0$, for all $\alpha$. 


\begin{definition}
\textit{A \textit{density operator} \cite{resource,nielsen}, described by a \textit{density matrix}, is a Hermitian operator ($\rho(t)=\rho(t)^{\dag}$), positive ($\rho(t)>0$) and with unit trace ($Tr[\rho(t)]=1$).
}
\end{definition}
\bl{We now analyze some properties} of the density operator and the corresponding matrix.
For a pure state, the density matrix is defined as a projector
\begin{equation}
\rho(t)=|\psi(t)\rangle\langle\psi(t)|,
\end{equation}
while for a mixed state we have
\begin{equation}
\rho(t)=\Sigma_ip_i|\psi_i(t)\rangle\langle\psi_i(t)|,
\end{equation}
where $p_i$ is the $i$-th weight, $p_i\in\mathbb{R}$, and $\Sigma_ip_i=1$.
The \textcolor{black}{index $i$ denotes the $i$-th robot. }
\bl{It can be easily observed from the above that a} pure state cannot be represented as the mixture of other states. Moreover, while a pure state is described by the wavefunction $|\psi(t)\rangle$, a mixed state is given by a superposition $\Sigma_ip_i|\psi_i(t)\rangle$.
Note that for a pure state, we have $\rho^2(t)=\rho(t)$ and $Tr[\rho^2(t)]=1$.

As a general note, to \bl{each} quantum state, \bl{one can determine} a corresponding density matrix, but given a density matrix, there \bl{can be more than a single quantum state} that corresponds to it, up to a phase difference \cite{nielsen}. In fact, the density matrix is \bl{only} a statistical object which captures the distribution of qubit information. \bl{The process of mapping a density matrix to a possible quantum state} is called \textit{quantum state tomography}, but it is a complex process that presents no particular advantage to us and therefore, irrelevant to this work ~\cite{nielsen}.
We present now the following remark that will be useful for our research.

\begin{remark}
\textit{ \bl{Two quantum states with similar density matrices are themselves similar, even though multiple states can correspond to a single density matrix. This similarity, or \emph{closeness} between the density matrices is quantified using a metric known as fidelity, with with values comprised between $0$ (low fidelity) and $1$ (high fidelity) \cite{nielsen}.}  A commonly used metric for this is the \textit{trace distance}, which provides information on how distinguishable two quantum states are. The trace distance between two density matrices $\rho_1(t)$ and $\rho_2(t)$ is defined as:
\begin{equation}
D[\rho_1(t),\rho_2(t)] = \frac{1}{2}Tr|\rho_1(t)-\rho_2(t)|,
\end{equation}
with 
\begin{equation}
|\rho_1(t)-\rho_2(t)| \coloneq \sqrt{(\rho_1(t)-\rho_2(t))^t(\rho_1(t)-\rho_2(t))}.
\end{equation}
}
\end{remark}

For a system that consists of two parts  (e.g. two \bl{swarm} particles, a particle and an agglomerate of particles, or two agglomerates of particles)
A and B, each of them described by density matrices $\rho_A(t)$ and $\rho_B(t)$ respectively, \bl{one can describe the density matrix of the system by}
\begin{equation}
\rho_{AB}(t)=\rho_A(t)\otimes\rho_B(t).
\end{equation}
To recover the information on state A from the matrix $\rho_{AB}(t)$, we need to compute its partial trace with respect to the state B, which is defined as
\begin{equation}
\begin{split}
Tr_B[\rho_{AB}(t)]& = \Sigma_B[\rho_{AB}(t)] \\ & = \Sigma_n\rho_A(t)\otimes \langle n|\rho_B(t) | j\rangle \\
&=\rho_A(t),
\end{split}
\end{equation}
where $n$ indicates the qubit which can be $0$ or $1$, and thus the sum involves the elements $\langle 0| ... |0\rangle$ and $\langle 1| ... |1\rangle$, that is, the measurement of the density operator along states $0$ and $1$, respectively.
The operation of partial trace is helpful, for example, to detach the information of a system from its environment. When we have a system composed by multiple elements, we can consider the sub-system of interest as \textit{the} system, and the other elements as part of the environment (as stated in the previous section). 

In Section \ref{new_model}, we use the idea of tensor product to build up a density matrix for the robotic swarm starting from individual density matrices, and the partial trace to achieve the inverse result, i.e., recovering information on single robots from the overall matrix. 


\section{\textcolor{black}{Quantum Robotic Swarm Modeling}}\label{single_robot_model}

\textcolor{black}{Before moving to the density matrix-based description of a quantum robotic swarm, we establish a correspondence between the parameters of the swarm by introducing the adopted bra-ket notation for a single robot.} \bl{To do so, similarly to~\cite{swarm_paper}}, we consider the position of each robot \ma{in the 2-dimensional plane as described by the superimposition of qubits along the $x$ and $y$ directions (as described in Section~\ref{sec:prelim}),}  and its success in target finding as probability amplitudes. \bl{Specifically, these amplitudes are used the construct the density matrices of the single robots.} Then, the density matrix of the swarm can be obtained via the tensor product of the individual density matrices.


Let us consider an $\mathcal{R}^2$ mobility plane ($x$ and $y$ axes), with a certain degree of success in reaching the target. Let $|\psi_{R_i}(t)\rangle$ be the state of robot $R_i$ at time $t$, defined as
\begin{equation}\label{caso_completo}
\begin{split}
|\psi_{R_i}(t)\rangle = &\alpha_i^1(t)|111\rangle + \alpha_i^2(t)|110\rangle + \alpha_i^3(t)|100\rangle + \\&\alpha_i^4(t)|000\rangle + \alpha_i^5(t)|001\rangle + \alpha_i^6(t)|010\rangle+
\\& \alpha_i^7(t)|101\rangle+
\alpha_i^8(t)|011\rangle.
\end{split}
\end{equation}
\bl{$|\psi_{R_i}(t)\rangle$ is the superimposition state vector of 3 qubits,} \textcolor{black}{of the form $|a,b,c\rangle$, with qubits $a,\,b,\,c$,} where the first qubit indicates the position projection along a segment $[0,1]$ on the x-axis ($0$ and $1$ are the two extremes), the second qubit denotes the projection of the position along a segment $[0,1]$ on the y-axis (similarly defined), \ma{and the third qubit indicates the success of finding the target, i.e., $|0\rangle$ indicates failure and $|1\rangle$ indicates success}; the coefficients $\alpha_i^j(t)$, with $i$ indicating the robot, and $j$ indicating the eigenvector, 
denote the probability amplitudes for the respective eigenstates/eigenvectors, satisfying $\sum_{j=1}^8|\alpha_i^j(t)|^2=1$.

\section{\textcolor{black}{Density Matrix-based Modeling of Quantum Robotic Swarm}}
\label{new_model}

\textcolor{black}{Considering the quantum state of a single robot, expressed in Eq. \eqref{caso_completo}, we can now propose the}
 density-matrix depiction of the swarm of robots. 
\bl{Then, for a robot $R_i$, the density operator is thus
\begin{equation}\label{density_Ri}
\rho_{R_i}(t)=|\psi_{Ri}(t)\rangle\langle\psi_{R_i}(t)|,
\end{equation}
and the state of $R_i$ can be written as a column matrix
\begin{equation}
|\psi_{R_i}(t)\rangle=\left(\begin{matrix} \alpha_i^1(t) \\ \alpha_i^2(t) \\ \dots \\ \alpha_i^8(t) \\ \end{matrix}  \right).    
\end{equation}
The density matrix is thus obtained via the product between the column matrix and the row matrix of complex conjugates coefficients, as follows:
\begin{equation}
\rho_{R_i}(t) = \left(\begin{matrix} \alpha_i^1(t)\\ \alpha_i^2(t) \\ \vdots \\ \alpha_i^8(t) \\ \end{matrix}  \right)\left(\alpha_i^{1,\ast}(t),\;\alpha_i^{2,\ast}(t),\; \dots,\;\alpha_i^{8,\ast}(t) \right).  
\end{equation}}

\begin{remark}
\textit{ \bl{In this article, we only} consider real values for the probability amplitudes, so the complex conjugates are equal to the original values. The time evolution of a single robot can be obtained via a suitable time-evolution operator acting on the corresponding density operator. Such an operator can include a dependence on the state of the other robots, acting according to the scope of the robotic mission. In \cite{swarm_paper}, the swarm behavior emerged from pairwise robotic interactions modeled upon  a quantum circuit. Here, we consider a more quantum-mechanics style approach. \textcolor{black}{Also, the matrix size does not increase when the swarm becomes bigger (with more units).}   
}
\end{remark}

Then, the density matrix of a swarm of $N$ robots is computed as
\begin{equation}
\rho_{swarm}(t)=\rho_{R_1}(t)\otimes...\otimes\rho_{R_N}(t),
\end{equation}
and the information on the $i^{\text{th}}$ robot can be retrieved performing the partial trace on all the other robots using

\bl{\begin{equation}
\rho_{R_i}(t) = Tr_{R_{j \neq i}(t)}(\rho_{swarm}(t)),
\end{equation}}
\bl{where $Tr_{R_{j \neq i}(t)}$ \textcolor{black}{denotes the sum of the diagonal elements of $\rho(t)$ with the exclusion of the ones indexed as $i$, that is, corresponding to the $i$-th robot.}}
\bl{The formulation of a quantum swarm in terms of tensor products of density matrices unfortunately has a major drawback: it results in large matrix sizes for large swarm sizes.
An alternate approach to alleviate this problem is to consider the swarm as a mixed state given by the superposition of pure states that represent single robots. This reduces the size of the matrix associated with the robot swarm, making the description more concise, compared to~\cite{swarm_paper}. Indeed, one can also consider single robots as mixed states, in which case, one needs to resort once again to consider the tensor product of density matrices for a clear representation of density matrix of the swarm.}


\bl{To illustrate this, consider an $N$-robot swarm, where the density matrix of the $i^{\text{th}}$ robot is given by $\rho_i(t) = |\psi_{R_i}(t)\rangle\langle\psi_{R_i}(t)|$. Considering the swarm as a \emph{mixed state}, its density matrix can be simply computed as}
\bl{\begin{equation}\label{mixture_abstract}
\rho_{swarm}(t) = \sum_{i=1}^N w_i \rho_i(t),
\end{equation}}
\bl{where $w_i$, with $i \in \{1,\ldots,N\}$ robots, are constant weights such that $\sum_i w_i = 1$.}
If the contribution of \bl{all the robots} \bl{is equal}, eq. \eqref{mixture_abstract} becomes
\begin{equation}
\begin{split}
\rho_{swarm}(t) &= \frac{1}{N} \sum_{i=1}^N \rho_i(t) = \\
&=\frac{1}{N} \sum_{i=1}^N |\psi_{R_i}(t)\rangle\langle\psi_{R_i}(t)| 
\end{split}
\end{equation}
Eq. \eqref{mixture_abstract} involves the sum of matrices, not their tensor product, and thus the dimension of the density matrix of the swarm, while seen as a mixed state, remains the same, as the dimensions of the single density matrices for the robots.

\bl{Moreover, while the swarm modeling approach presented in~\cite{swarm_paper} utilizes interaction terms in the swarm submatrices to describe the interactions between robots, in this work, we neglect the interaction terms by considering that each robot sends all information to all the other robots, and each robot receives information from all the other robots. Thus, copying this information in the interaction sub-matrices would just be redundant. } Note that the effect of ``interaction'' are automatically included in the density-matrix depiction of the swarm, seen as a whole system in a mixed state, resulting from the mixing of single states of the single robots. \bl{However, the utilization of explicit interation terms largely depends on the size of the considered devices: while bigger robots may use classical wireless communications, for micro/nano scale devices, newer opportunities arise \cite{wireless03}, since they are nearer to quantum systems \cite{wireless04}.}




In the following section, we illustrate our proposed method with several examples.

\section{Examples}\label{examples}



\subsection{Pure States}

As a first example, we consider a toy-swarm of two robots, $R_1$ and $R_2$, moving along a line of length $1$, i.e., from $x=0$ to $x=1$. Suppose that the target is located in $x=1$.
From Eq. \eqref{caso_completo}, we can write
\begin{equation}\label{state1}
|\psi_{R_i}(t)\rangle = \alpha_i^1(t)|11\rangle + 
\alpha_i^2(t)|10\rangle + 
\alpha_i^3(t)|01\rangle + 
\alpha_i^4(t)|00\rangle,
\end{equation}
where the first qubit indicates the position along the $x$ axis, and the second qubit the degree of success in target finding, and $\sum_j^4|\alpha_i^j(t)|^2=1$.
\bl{With a slight abuse of notation, we represent the position of the robot instead with the peak amplitude of the wavefunction of the position~\cite{swarm_paper}. Moreover, we remove the time argument from the state for notational simplicity by assuming that the positions are considered at a pre-defined time instant.} Now suppose that $R_1$ is at $x=0.5$ (the peak of the wavefunction is centered in $0.5$), and it has $0$ success (the peak of the ``success'' wavefunction is in $0$). On the other hand, $R_2$ at $x=1$ and with a successful result in target finding ($1$); it is thus in an eigenstate. Substituting the non-null coefficients $\alpha^2_1=\alpha^4_1=\sqrt{1/2}$ and $\alpha^1_2=1$ for the two robots, respectively, we obtain
\begin{align}
|\psi_1\rangle & =\frac{1}{\sqrt{2}}(|10\rangle + |00\rangle),\, \\
|\psi_2\rangle & =|11\rangle,
\end{align}
Correspondingly, the density matrix of $R_1$ is given by
\begin{equation}\label{etichetta_rho1}
\rho_1=|\psi_1\rangle\langle\psi_1|=\frac{1}{2}(|10\rangle\langle 10|
+
|10\rangle\langle 00|
+
|00\rangle\langle 10|
+
|00\rangle\langle 00|
),
\end{equation}
and that of $R_2$ is given by
\begin{equation}\label{etichetta_rho2}
\rho_2=|\psi_2\rangle\langle\psi_2|=|11\rangle\langle 11|.
\end{equation}
\textcolor{black}{We recall the definition of the density matrix $\rho_j$, computed from the ket $|\psi_j\rangle$ and the bra $\langle\psi_j|$. A ket lives in a Hilbert space $\mathcal{H}$, the bra in its dual space (that, for finite-dimensional Hilbert spaces, is its conjugate transpose). The density operator acts on the Hilbert space, and the density matrix is its representation in terms of matrix. As a general note, in this article, we adopt the term ``matrix'' to denote both the operator and it matrix representation. The density operator lives in the space of bounded linear operators acting on the Hilbert space: $\rho_j\in\mathcal{B}(\mathcal{H})$. To obtain the matrix representation, we have to consider, as rows, the kets $|11\rangle,\,|10\rangle,\,|01\rangle,|00\rangle$, and as columns, the bras $\langle11|,\,\langle10|,\,\langle01|,\langle00|$. In this way, for the density matrices of Eq. \eqref{etichetta_rho1} and \eqref{etichetta_rho1}, we can obtain the following matrix forms:}
\begin{equation*}
\rho_1=\frac{1}{2}
\left(
\begin{matrix}
0 & 0 & 0 & 0 \\
0 & 1 & 0 & 1 \\
0 & 0 & 0 & 0 \\
0 & 1 & 0 & 1 \\
\end{matrix}
\right),\,
\rho_2=
\left(
\begin{matrix}
1 & 0 & 0 & 0 \\
0 & 0 & 0 & 0 \\
0 & 0 & 0 & 0 \\
0 & 0 & 0 & 0 \\
\end{matrix}
\right).
\end{equation*}
The density matrix of the toy swarm is obtained as their tensor product.
If we have the state as the initial information, and we want to recover the probability of the robot to be at $x=1$, we compute:
\begin{equation}\label{param}
|\langle1_{x}|\psi_1\rangle|^2=0.5,\,|\langle1_{x}|\psi_2\rangle|^2=1.
\end{equation}
The probability amplitude that the swarm barycenter is in $x=1$ is the mean of the two terms in Eq. \eqref{param}, i.e., $0.75$.
An ideal swarm with two robots, both centered on the target in $1$ and successful, of course has a barycenter with probability $1$ to be found on $1$. A measure of distance between the swarm barycenter and the ideal-swarm barycenter is the difference $|0.75-1|$. 
The information on positions of single robots can also be retrieved from the density matrix; this is useful in cases where we only have access to the density matrix. The partial trace of the individual density matrices with respect to the second qubit (taking away the information on ``success'', i.e., $s$, \textcolor{black}{to not be confused with the initial for \textit{swarm}}) provides information on the individual robot's position.
Concerning $R_1$, after a few passages we obtain:
\begin{equation}\label{reduced1}
\begin{split}
&Tr_{s}\rho_1=\langle1_{s}|\rho_1|1_{s}\rangle+\langle0_{s}|\rho_1|0_{s}\rangle=\\
&=\frac{1}{2}(|1\rangle\langle 0|+|0\rangle\langle 0|),
\end{split}
\end{equation}
while for $R_2$ we simply have
\begin{equation}\label{reduced2}
Tr_{s}\rho_2=\langle1_{s}|\rho_2|1_{s}\rangle+\langle0_{s}|\rho_2|0_{s}\rangle=|1\rangle\langle1|.
\end{equation}
We notice that $R_1$ has $1/2$ probability of being in $1$, while $R_2$ has probability $1$. Thus, the barycenter of the swarm is their mean, $0.75$. In this way, we prove that the same information can be recovered from density matrices. Tracing the overall density matrix $\rho_{swarm}=\rho_{12}=\rho_1\otimes\rho_2$ with respect to the second qubit, we can also recover the position of the swarm barycenter. The same applies to the ideal swarm centered on the target. The comparison between the values for the given swarm and the ideal one, gives a measure of the precision with which the target has been reached.
We can define an opportune distance between matrices as the trace of the difference between the reduced matrices, that is $Tr(|\langle1_s|\rho_{12}|1_s\rangle|-|\langle1_s|\rho_{12(T)}|1_s\rangle|)$, where we focus on the probability of finding the barycenter close to $x=1$, and where $\rho_{12}$ is the density matrix of the ideal swarm. It is clear that we can think of a whole class of ``successful'' swarms, whose robots are inside a small neighbor of the target. \bl{In addition, using our modeling framework, one may also consider stability problems. Specifically, one may define a Lyapunov as the Hamiltonian of the barycenter of the swarm, which is a function of the robot's position displacements (potential term) and momentum (speed term). Stability will be investigated in future research.}

\subsection{Mixed States}
\bl{Consider again the 2-robot swarm described above. As discussed previously, one can significantly benefit in formalizing the robot swarm as a mixed state consisting of robots that are described as pure states. In this case, the density matrix of the swarm would be defined as the weighted sum of the states, rather than as their tensor product.}
\begin{equation}\label{mixed}
\rho_{swarm} = \frac{1} 
{2}|\psi_1\rangle\langle\psi_1| + \frac{1}{2}|\psi_2\rangle\langle\psi_2|
\end{equation}
The information on a single robot can still be recovered via the partial trace, tracing the $\rho_{swarm}$ with respect to the other robot(s). In matrix form, Eq. (\ref{mixed}) would give
\begin{equation}
\rho_{swarm}=
\left(
\begin{matrix}
1/2 & 0 & 0 & 0 \\
0 & 1/4 & 0 & 1/4 \\
0 & 0 & 0 & 0 \\
0 & 1/4 & 0 & 1/4 \\
\end{matrix}
\right).
\end{equation}
We verify that $Tr(\rho_{swarm})=1$; to check that $\rho_{swarm}$ is a mixed state, we compute the trace of the squared matrix, verifying that it is smaller than 1
\begin{equation*}
Tr(\rho_{swarm}^2)=
Tr\left(
\begin{matrix}
1/4 & 0 & 0 & 0 \\
0 & 1/8 & 0 & 1/8 \\
0 & 0 & 0 & 0 \\
0 & 1/8 & 0 & 1/8 \\
\end{matrix}
\right)=\frac{1}{2}.
\end{equation*}
For the swarm we \textcolor{black}{compute the overall energy $E$ as} $E=\Sigma_{i=1}^NE_i$, with $N=2$ in our example, and $E_i$ the energy of each single robot. Then, we compute $\Delta E= E(t)-E(t')$, where the time dependence is related with the instantaneous peak position, and we can finally choose the Hamiltonian as

\begin{equation}
H=\frac{\Delta E}{2}\sigma_z=\frac{\Delta E}{2}(|0\rangle\langle0|-|1\rangle\langle1|).
\end{equation}
To reduce the dimensions of the problem, we can perform the partial trace operation on $\rho_{swarm}$ with respect to the \textit{success} information, \textcolor{black}{that we can denote with $s$ as in the previous section.} We also perform this operation beforehand on the density matrices of single robots, as showed in Eq. (\ref{reduced1}, \ref{reduced2}). The reduced density matrix of the swarm as a mixed state is  \textcolor{black}{obtained by tracing on the success $s$}
\begin{equation}
\begin{split}
\rho_{swarm}^{r}&=\frac{1}{2}Tr_s(\rho_1)+\frac{1}{2}Tr_s(\rho_2)=\\&=\frac{1}{4}(|1\rangle\langle0|+|0\rangle\langle0|+2|1\rangle\langle1|).
\end{split}
\end{equation}
If we consider the ideal situation where the interaction of the swarm with the environment is absent or negligible, the $L$ operators of the master equation in Lindblad form are negligible as well, and the master equation is reduced to
\begin{equation}\label{first_term}
\mathcal{L}(\rho_{swarm}^r)=-i[H,\rho_{swarm}^r]=-i\frac{\Delta E}{2}(|0\rangle\langle0|-|1\rangle\langle1|).
\end{equation}
To consider the interaction with the environment, we have to include non-null $L$ operators. For the sake of simplicity, we can choose $L=|0\rangle\langle0|$.
Thus, the remaining terms of the master equation are computed as
\begin{equation}\label{other_terms}
\begin{split}
L^T\rho_{swarm}^r L^{\ast} =&\frac{1}{4}|0\rangle\langle 0|
\\
-\frac{1}{2}L^{\dag}L\rho_{swarm}^r =&
-\frac{1}{8}|0\rangle\langle0|\\
-\frac{1}{2}\rho_{swarm}^r L^{\dag}L =&-\frac{1}{8}(|1\rangle\langle0|+|0\rangle\langle0|).
\end{split}
\end{equation}

Summing up the terms of Eqq. (\ref{first_term},\ref{other_terms}), we obtain
\begin{equation}
\mathcal{L}(\rho_{swarm}^r)=-i\frac{\Delta E}{2}(|0\rangle\langle0|-|1\rangle\langle1|)-\frac{1}{8}|1\rangle\langle0|.
\end{equation}

\subsection{Other Computational Examples}

\begin{figure}
\centering
\includegraphics[width=0.9\columnwidth]{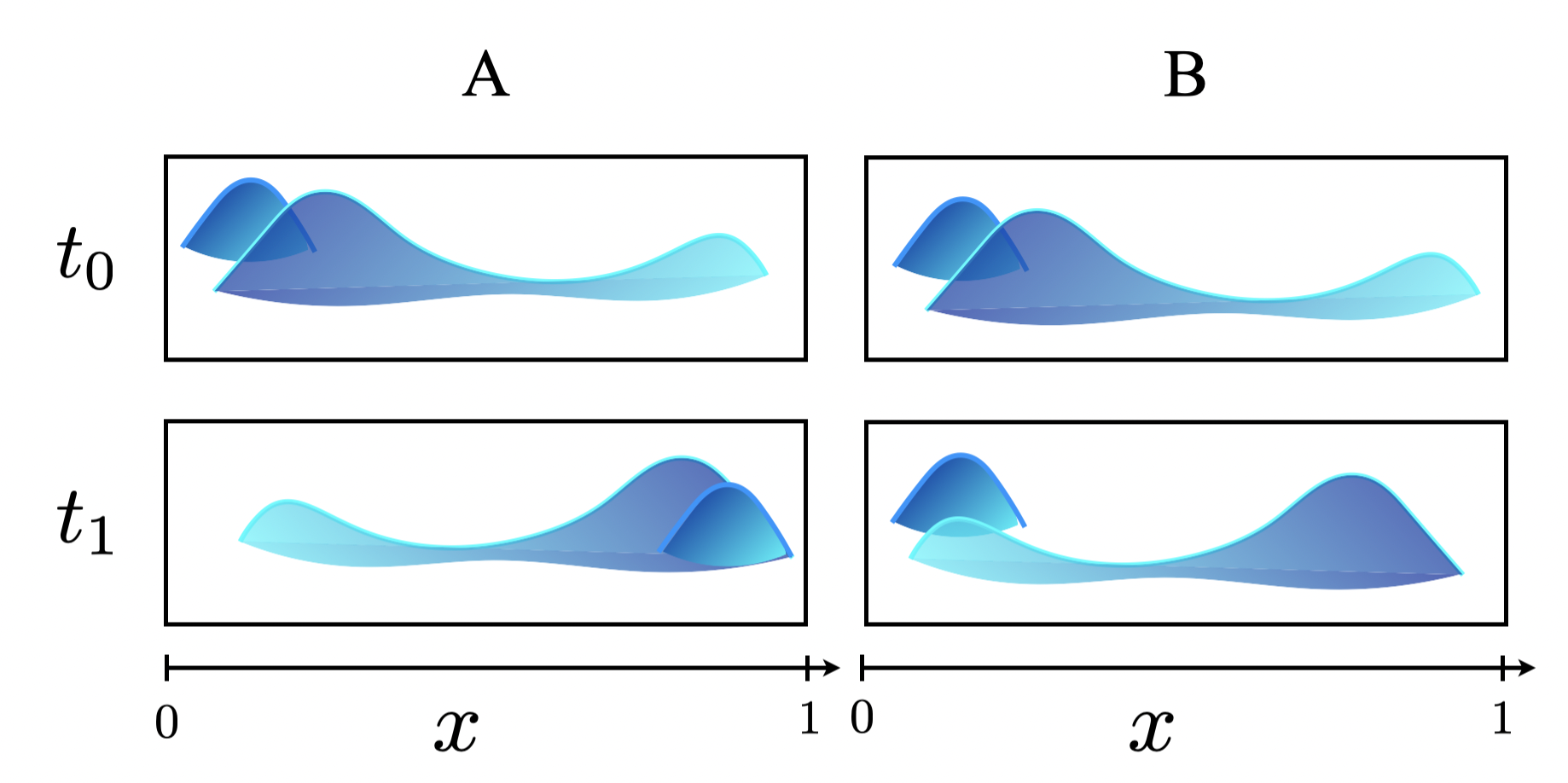}
\caption{Quantum-like representation of a toy 2-robot swarm, where the position of the robot is considered as the weighted peak of the corresponding wavefunction.}\label{cases_A_B_b}
\end{figure}

\ma{Next,} consider another toy swarm, constituted by two robotic fish $R_1,\,R_2$, moving along the segment $[0,1]$ of the $x$ axis, as shown in Figure \ref{cases_A_B_b}. For the sake of simplicity, we neglect here information about success in target reaching, focusing only on information about position.
Case A presents two robots maintaining a constant distance between them, while case B presents two robots getting distant from each other.

Let us focus on case A, an example of swarm compactness over time.
Figure \ref{cases_A_B_b} exemplifies the passage to a quantum representation, where the positions are substituted by the peaks of the wavefunctions. \textcolor{black}{As an example, we choose to write} the density operator of the swarm as a  mixed state:
\begin{equation}
\rho^A(t_0)=\frac{1}{2}|0\rangle\langle0|+\frac{1}{2}(\sqrt{0.8}|0\rangle+\sqrt{0.2}|1\rangle)(\sqrt{0.8}\langle0|+\sqrt{0.2}\langle1|).
\end{equation}

The swarm at the following time point is represented by the density operator:
\begin{equation}
\rho^A(t_1)=\frac{1}{2}|1\rangle\langle1|+\frac{1}{2}(\sqrt{0.8}|1\rangle+\sqrt{0.2}|0\rangle)(\sqrt{0.8}\langle1|+\sqrt{0.2}\langle0|).
\end{equation}
The respective matrices, for the swarm at $t_0$ and at $t_1$, are:
\begin{equation}
\rho^A(t_0)=\left(
\begin{matrix}
0.9 & 0.2 \\
0.2 & 0.1
\end{matrix}
\right),\,\,\,\,
\rho^A(t_1)=\left(
\begin{matrix}
0.1 & 0.2 \\
0.2 & 0.9
\end{matrix}
\right).
\end{equation}
\textcolor{black}{To obtain these results, we have to multiply bra-kets, and sum the terms:
\begin{equation}
\begin{split}
\rho^A(t_0)=\frac{1}{2}|0\rangle\langle0|+\frac{1}{2}(\sqrt{0.8}|0\rangle+\sqrt{0.2}|1\rangle)(\sqrt{0.8}\langle0|+\sqrt{0.2}\langle1|) = \\
\frac{1}{2}|0\rangle\langle0| + \left(\frac{0.8}{2}|0\rangle\langle0|+\frac{\sqrt{1.6}}{2}|0\rangle\langle1|+\frac{\sqrt{1.6}}{2}|1\rangle\langle0|+\frac{0.2}{0}|1\rangle\langle1|\right) =\\ 0.9|0\rangle\langle0|+0.2|0\rangle\langle1|+0.2|1\rangle\langle0|+0.1|1\rangle\langle1|,
\end{split}
\end{equation}
and then, organizing the coefficients $0.9,\,0.2,\,0.2,\,0.1$ inside the matrix with columns $|0\rangle,\,|1\rangle$ and rows $\langle0|,\,\langle1|$, we get the result.}

\textcolor{black}{Supposing that the variation of energy is positive}, we obtain:
\begin{equation}\label{condition_stab}
Tr[\mathcal{L}(\rho^A_{t_1})\Delta\rho^A] = -0.16 i\Delta E <0.
\end{equation}
\textcolor{black}{Recalling the definition of $\rho$-based stability in \cite{emzir}, in the case of positive $\Delta E$, the condition of Eq. \eqref{condition_stab} denotes stability for the case $A$. In this article, we chose to just present a general approach to a quantum swarm, but this formalism can be helpful for future technological developments of the research.}

We conclude the section of Examples with a case of target-reaching application.
Let us start from the 
previously-found swarm density matrix, \textcolor{black}{that we will simply indicate with $\rho_{swarm}$}
\begin{equation}
\rho_{\textcolor{black}{swarm}}=\left(
\begin{matrix}
0.9 & 0.2 \\
0.2 & 0.1
\end{matrix}
\right),
\end{equation}
and let us consider a target in $|0\rangle\langle0|$, as an ideal swarm centered on the target, described by a state $\rho_{\textcolor{black}{T}}$ having density matrix
\begin{equation}
\rho_T=\left(
\begin{matrix}
1 & 0 \\
0 & 0
\end{matrix}
\right).
\end{equation}
The distance between the swarm and the target can be computed 
with the trace distance definition:
\begin{equation}
\begin{split}
d(\rho_{swarm}, \rho_T)=
\frac{1}{2}Tr[\sqrt{(\rho_{swarm}, \rho_T)^t(\rho_{swarm}, \rho_T)}] = 0.05.
\end{split}
\end{equation}

\textcolor{black}{The adoption of density matrices for swarms of quantum robots can pave the way toward applications in control theory, adapting to our cases criteria of stability for open quantum systems, that are based on density matrices and suitable choices of Lyapunov functions \cite{emzir}.}

\subsection{\textcolor{black}{Movements on a 2D surface, First Examples with Qiskit, and Larger Swarms}}\label{example_qiskit}

\begin{figure}[ht!]
\centering
\includegraphics[width=\textwidth]{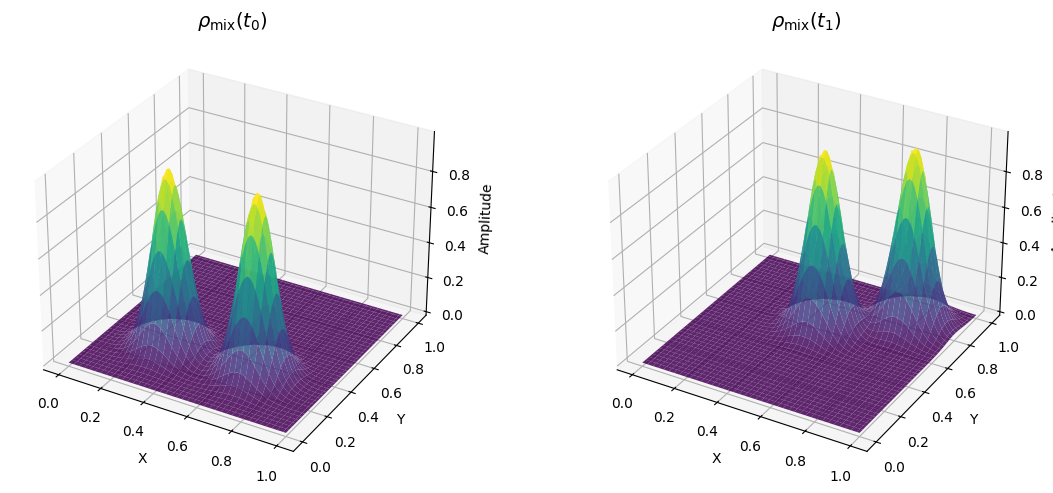}
\caption{\ma{Wavefunction} visualization of the probability amplitudes to find the two robots-particles described in Eq. \eqref{density_matrix_initial_state_working} and \eqref{density_matrix_final_state_working}, respectively.}\label{3D}
\end{figure}

\ma{First,} we propose a working example to visualize a toy swarm of two robots as surfaces representing the wavefunctions, \textcolor{black}{neglecting the target, and instead focusing only on the swarm.} 
\bl{The process describes a global-local approach, wherein density matrices of the swarm and a time evolution operator are computed to determine the evolution of the swarm. Then, local distribution of the individual robots are updated accordingly. Such a process, as opposed to the local-to-global approach proposed in \cite{swarm_paper}, allows to streamline local information of the robots from global information on the swarm.}
Suppose that the initial state of particle 1 at time $t_0$ is
\begin{equation}
|\psi_1(t_0)\rangle = \frac{1}{\sqrt{2}}(|00\rangle + |11\rangle),
\end{equation}
and the initial state of particle 2 is
\begin{equation}
|\psi_2(t_0)\rangle = |00\rangle
\end{equation}
Thus, the density matrix for the first particle-robot is
\begin{equation}
\rho_1(t_0) = \frac{1}{2}(|00\rangle\langle00|+|11\rangle\langle11|)=\frac{1}{2}\left( \begin{matrix} 1 & 0 & 0 & 1 \\
0 & 0 & 0 & 0 \\
0 & 0 & 0 & 0 \\
1 & 0 & 0 & 1
\end{matrix}\right),    
\end{equation}
and the density matrix for the second particle-robot is
\begin{equation}
\rho_2(t_0) = |00\rangle\langle00|=\left( \begin{matrix}
1 & 0 & 0 & 0 \\
0 & 0 & 0 & 0 \\
0 & 0 & 0 & 0 \\
0 & 0 & 0 & 0
\end{matrix}\right).   
\end{equation}
From this, we \textcolor{black}{obtain the density matrix of the swarm as a mixed density matrix. To highlight this, here we use the pedex \textit{mix}:}
\begin{equation}\label{density_matrix_initial_state_working}
\rho_{mix}(t_0)=\frac{\rho_1(t_0) + \rho_2(t_0)}{2} = 
\left(\begin{matrix}
0.75 & 0. &  0. &  0.25 \\
0. &  0. &  0. &  0. \\
0. &  0. &  0. &  0. \\
0.25 & 0. &  0. &  0.25
 \end{matrix}
 \right)
\end{equation}
We choose the final state of particle 1 at $t_1$ as
\begin{equation}
|\psi_1(t_1)\rangle = \frac{\sqrt{0.2}}{\sqrt{0.8^2 + 0.2^2}}|00\rangle + \frac{\sqrt{0.8}}{\sqrt{0.8^2 + 0.2^2}}|11\rangle,
\end{equation}
and that of particle 2 as
\begin{equation}
|\psi_2(t_1)\rangle = \frac{\sqrt{0.1}}{\sqrt{0.9^2 + 0.1^2}}|00\rangle + \frac{\sqrt{0.9}}{\sqrt{0.9^2 + 0.1^2}}|11\rangle.
\end{equation}
Finally, the density matrix of the mixed state is given by
\begin{equation}\label{density_matrix_final_state_working}
\rho_{mix}(t_1)=\left(\begin{matrix}
0.15 & 0. &  0. &  0.35 \\
0. &  0. &  0. &  0. \\
0. &  0. &  0. &  0.  \\
0.35 &  0. &  0. &  0.85
\end{matrix}\right).
\end{equation}
Figure \ref{3D} shows the robots corresponding to the two ``particles'' as \textcolor{black}{wavefunctions}, whose height along the $z$-axis denotes the probability amplitude. Such a representation is an additional overcoming of the visualizations of \cite{swarm_paper}, where a classic position was considered in correspondence to the peak of the wavefunctions. 
\begin{algorithm}[ht!]
\begin{algorithmic}
\caption{computation of density matrices}
\label{algo1}


\State \textbf{Input:} $N$ robots with states $\{\psi_i\}_{i=1}^N$, 
$N$ of robots, optional: target position as $|\psi_T\rangle$, maximum number of time steps $n$ to reach the target

\State \textbf{Output:}
swarm density matrix at each time step $t$; visualizations


\State \textbf{Procedure:}
initialize robot states $\psi_i(t_0)$
\If{target is present}
\State compute density matrix of the target as $\rho_T = |\psi_T\rangle\langle\psi_T|$
\EndIf
\For{each time step}
\State update individual robot states according to swarm interaction rules
\State compute density matrices for the individual robots as $\rho_i(t) = |\psi_i(t)\rangle\langle\psi_i(t)|$
\State compute swarm density matrix $\rho_{swarm}(t)$ as $\frac{1}{N}\sum_i^N\rho_i(t)$
\If{target is present}
\State compute trace distance between the matrix of the swarm and the matrix of the target as $tr|\rho_{swarm}(t)-\rho_T(t)|$
\EndIf
\EndFor

\end{algorithmic}
\end{algorithm}

\textcolor{black}{The computation of density matrices can be automatized by using Qiskit with the help of loading functions \textit{Statevector} and \textit{DensityMatrix}. The matrices and the visualizations in Figure \ref{3D} are obtained via the implementation in Python (code provided in Code Availability) of  Algorithm \ref{algo1}. For our implementation, we chose a  Gaussian appearance for the wavefunctions.}
\textcolor{black}{The code yields Figure~\ref{3D}, and the robot states as well as density matrices for each robot as well as for the whole swarm at $t_0$ obtained as:}
\ma{
\begin{align}
|\psi_1(t_0) \rangle = [0.707, 0, 0, 0.707],  \quad
|\psi_2(t_0) \rangle = [1, 0, 0, 0] \label{eq:initial}
\end{align}
\begin{align}
\rho_1(t_0) = \left( \begin{matrix}
    0.5 & 0 & 0 & 0.5 \\
0 & 0 & 0 & 0 \\
0 & 0 & 0 & 0 \\
0.5 & 0 & 0 & 0.5
\end{matrix} \right), \ 
\rho_1(t_0) = \left( \begin{matrix}
    1 & 0 & 0 & 0 \\
0 & 0 & 0 & 0 \\
0 & 0 & 0 & 0 \\
0 & 0 & 0 & 0
\end{matrix} \right), \ 
\rho_{swarm}(t_0) = \left( \begin{matrix}
    0.75 & 0 & 0 & 0.25 \\
0 & 0 & 0 & 0 \\
0 & 0 & 0 & 0 \\
0.25 & 0 & 0 & 0.25
\end{matrix} \right)
\end{align}
while the final states and matrices for time $t_1$ is obtained as:
\begin{align}
|\psi_1(t_1) \rangle = [0.447, 0, 0, 0.894]  \quad
|\psi_2(t_1) \rangle = [0.316, 0, 0, 0.949] \label{eq:initial}
\end{align}
\begin{align}
\rho_1(t_1) = \left( \begin{matrix}
    0.2 & 0 & 0 & 0.4 \\
0 & 0 & 0 & 0 \\
0 & 0 & 0 & 0 \\
0.4 & 0 & 0 & 0.8
\end{matrix} \right), \ 
\rho_1(t_1) = \left( \begin{matrix}
    0.1 & 0 & 0 & 0.3 \\
0 & 0 & 0 & 0 \\
0 & 0 & 0 & 0 \\
0.3 & 0 & 0 & 0.9
\end{matrix} \right), \ 
\rho_{swarm}(t_1) = \left( \begin{matrix}
    0.15 & 0 & 0 & 0.35 \\
0 & 0 & 0 & 0 \\
0 & 0 & 0 & 0 \\
0.35 & 0 & 0 & 0.85
\end{matrix} \right)
\end{align}}









\begin{figure}[ht!]
\centering
\includegraphics[width=\linewidth]{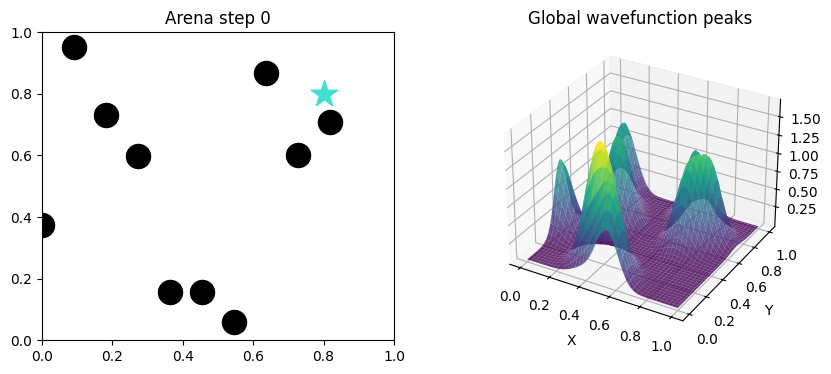}
\caption{Visualization of the 10-robot swarm at the beginning of our simulation. The turquoise star indicates the target, adopting the style of 2D representation of \cite{swarm_paper}.}
\label{10_robot_simulation_initial}
\end{figure}

\textcolor{black}{Now, we include visualizations for larger swarms. Here, we include a fixed target, and assume that the robot dynamics are such that they advance toward the target over time. This results in $3D$ representations of the swarm, with success of reaching the target included as in~\eqref{caso_completo}. Visualizations are fixed for a maximum number of time steps. In Figure \ref{10_robot_simulation_initial}, we show an example of a swarm of $N=10$ robots, whose initial states are defined by Eq. \eqref{ten_robots}:
\begin{equation}\label{ten_robots}
\begin{split}
&|\psi_1\rangle = \left[\frac{1}{\sqrt{2}}, 0, 0, 1/\sqrt(2)\right],\,\,\,\,
|\psi_2\rangle =
    \left[1, 0, 0, 0]\right],\\
&|\psi_3\rangle =
    \left[0, 0, -1, 0\right],\,\,\,\,
|\psi_4\rangle =
    \left[0, 1, 0, 0\right],\\
&|\psi_5\rangle =
    \left[0, \frac{1}{\sqrt{2}}, \frac{1}{\sqrt{2}}, 0\right],\,\,\,\,
|\psi_6\rangle =
    \left[\frac{1}{\sqrt{3}}, \frac{1}{\sqrt{3}}, 0, \frac{1}{\sqrt{3}}\right],\\
&|\psi_7\rangle =
    \left[-\frac{1}{\sqrt{3}}, \frac{1}{\sqrt{3}}, 0, \frac{1}{\sqrt{\sqrt{3}}}\right],\,\,\,\,
|\psi_8\rangle =
    \left[-\frac{1}{\sqrt{3}}, 0, 0, \frac{1}{\sqrt{3}}\right],\\
&|\psi_9\rangle =
    \left[-\frac{1}{\sqrt{2}}, 0, 0, -\frac{1}{\sqrt{2}}\right],\,\,\,\,
|\psi_{10}\rangle =
    \left[0,0,\frac{1}{\sqrt{2}}, \frac{1}{\sqrt{2}}\right].\\ 
\end{split}
\end{equation}
}

\textcolor{black}{With the help of Algorithm~\ref{algo1}, we obtain the density matrix of the swarm at the initial time $t_0$ as:
\begin{equation}\label{swarm_10_initial}
\rho_{swarm, 10}(t_0)=\left(
\begin{matrix}
0.3 & 0.067 & 0. &  0.067 \\
0.067 & 0.217 & 0.05 & 0. \\
0. & 0.05 & 0.2 & 0.05 \\
0.067 & 0. & 0.05 & 0.25 \\
\end{matrix}\right). 
\end{equation}
Then, we can define a fixed target that does not change over time with density matrix $\rho_T = |11\rangle\langle11|$, from 
$\psi_T = [0,0,0,1]$. Thus, the trace distance between the density matrix $\rho_{swarm, 10}(t_i)$ of our 10-robot swarm and $\rho_T(t_i)$ is $0.7402$. At this point, for the sake of simplicity, we consider a progressive movement of robots toward the target, for a given number of steps. Thus, we can compute the density matrix of the swarm at each time step, and we add the visualization of the classical peaks within an arena, as the 2D representation adopted in \cite{swarm_paper}.}

\begin{figure}[ht!]
\centering
\includegraphics[width=\linewidth]{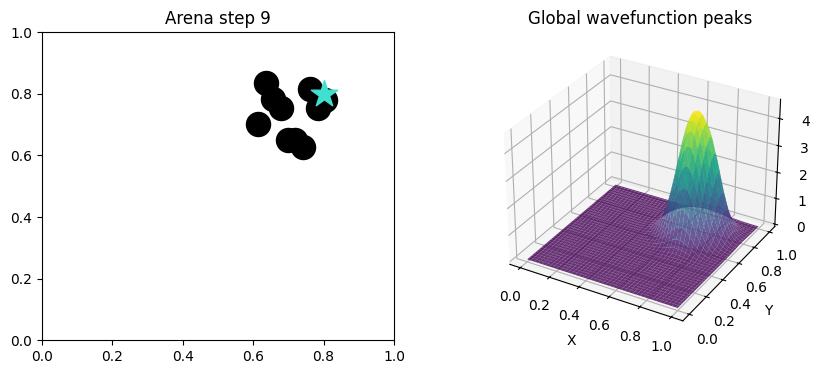}
\caption{Visualization of the 10-robot swarm once it reached the target.}
\label{10_robot_simulation_final}
\end{figure}

\begin{figure}[ht!]
\centering
\includegraphics[width=0.8\linewidth]{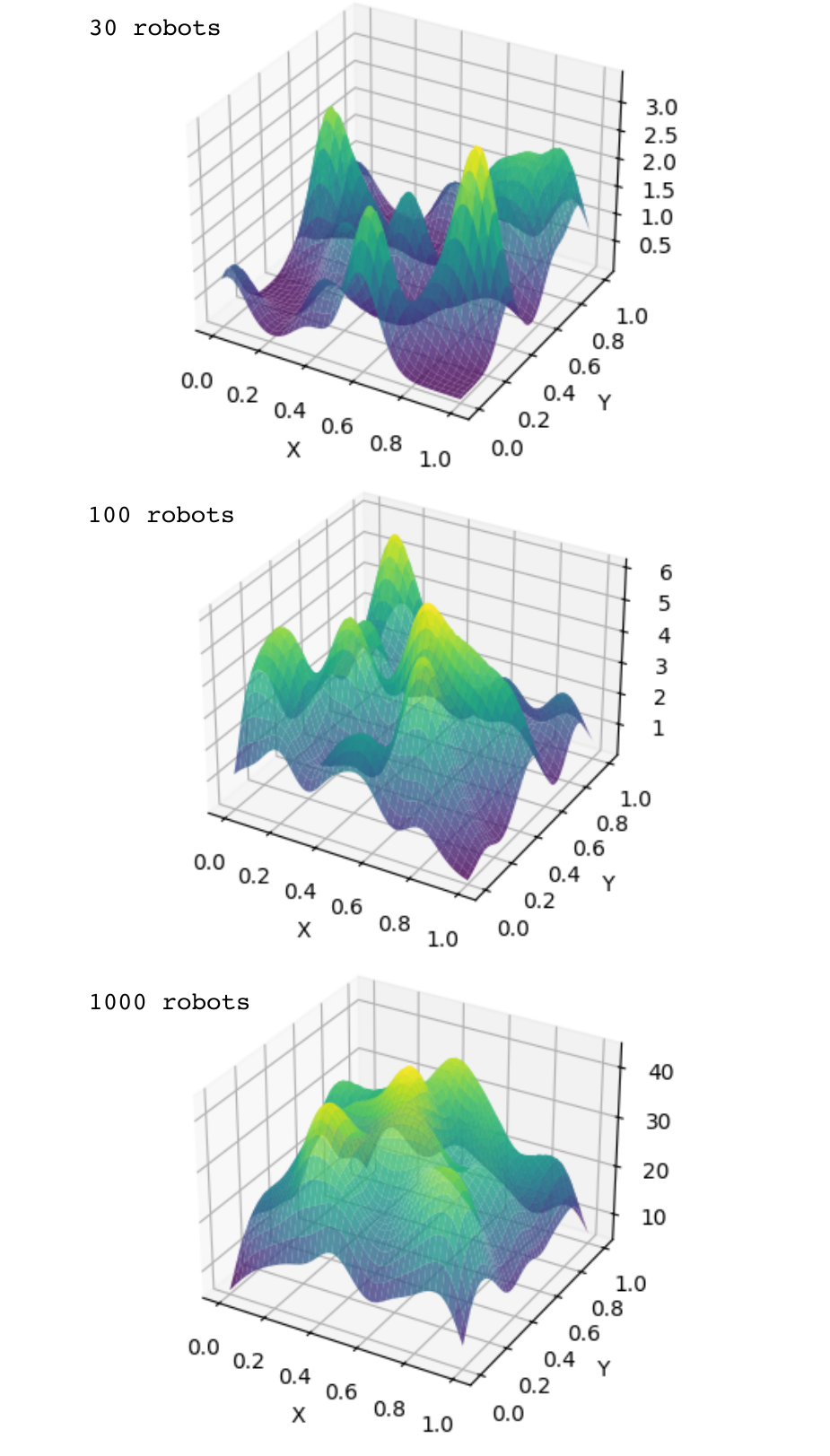}
\caption{Visualizations and density matrices of swarms of different sizes ($N=30,\,100,\,1000$ robots).}
\label{comparison_30-100-1000}
\end{figure}

\textcolor{black}{After $10$ steps, we get the following visualizations and swarm density matrix as:
\begin{equation}\label{swarm_10_final}
\rho_{swarm, 10}(t_1)=\left(
\begin{matrix}
0.241 & 0.032 & 0. & 0.074 \\
0.032 & 0.147 & 0.038 & 0.072 \\
0. & 0.038 & 0.137 & 0.029 \\
0.074 & 0.072 & 0.029 & 0.475\\
\end{matrix}\right). 
\end{equation}
At this step, when the swarm reaches the target, the trace distance between the density matrix of the swarm and the target is $0.5399$.
Figure \ref{10_robot_simulation_final} shows the final appearance of the swarm.}
\textcolor{black}{The initial positions of the wavefunctions peaks can also be randomized. Having applied these functions, we provide here an example of visualization with 30, 100, and 1000 robots, also showing their respective density matrices, see Figure \ref{comparison_30-100-1000}.}
\textcolor{black}{The matrices are given as:
\begin{equation}\label{larger_swarms}
\begin{split}
&\rho_{swarm,30}= \left(
\begin{matrix}
0.186 & 0.133 + 0.093i & 0.143 + 0.034i & 0.156 + 0.087i \\
0.133 - 0.093i & 0.276 & 0.191 - 0.076i & 0.239 + 0.002i \\
0.143 - 0.034i & 0.191 + 0.076i & 0.246 & 0.179 + 0.053i \\
0.156 - 0.087i & 0.239 - 0.002i & 0.179 - 0.053i & 0.291
\end{matrix}
\right)\\
&\rho_{swarm,100} = \left(
\begin{matrix}
0.237 & 0.174 + 0.009i & 0.153 - 0.007i & 0.183 \\
0.174 - 0.009i & 0.258 & 0.179 - 0.022i & 0.207 - 0.007i \\
0.153 + 0.007i & 0.179 + 0.022i & 0.227 & 0.188 + 0.014i \\
0.183 & 0.207 + 0.007i & 0.188 - 0.014i & 0.278
\end{matrix}\right)\\
&\rho_{swarm,1000} = \left(
\begin{matrix}
0.251 & 0.183 - 0.009i & 0.180 - 0.001i & 0.182 + 0.002i \\
0.183 + 0.009i & 0.253 & 0.180 + 0.001i & 0.182 + 0.008i \\
0.180 + 0.001i & 0.180 - 0.001i & 0.246 & 0.179 + 0.008i \\
0.182 - 0.002i & 0.182 - 0.008i & 0.179 - 0.008i & 0.251
\end{matrix}\right).\\
\end{split}
\end{equation}
}



\textcolor{black}{One can notice that there are complex terms the off-diagonal terms of the density matrices: these are the coherences, and they indicate \emph{entanglement} in the swarm. We can interpret them as the emerging of quantum effects from our definition of the swarm. However, a detailed investigation of this point is out of the scope of this article.}

\textcolor{black}{As it can be noticed from Eq. \ref{larger_swarms}, the size of the density matrices is not changed by the increasing of the size of the swarms, allowing for a more compact and less computationally challenging description of the swarm. In fact, in our representation, the size of the matrix describing the swarm depends only upon the degrees of freedom of its components, and their proximity to the target.}

\subsection{Time Evolution}\label{time_evolution}

\textcolor{black}{We are also interested in finding a method to approximate the time evolution operator, that is, a unitary matrix $O(t)$ such that:}
\begin{equation}
O(t_1)\rho_{\textcolor{black}{swarm}}(t_0) = \rho_{\textcolor{black}{swarm}}(t_1).
\end{equation}
\textcolor{black}{A possible technique is the singular value decomposition (SVD). We first compute the SVD of $\rho_{\textcolor{black}{swarm}}(t_0)$ obtaining two matrices $U$ and $V^{\dag}$ and then the SVD of $\rho_{\textcolor{black}{swarm}}(t_1)$  obtaining two other matrices $U'$ and $V'^{\dag}$, such as:}
\begin{equation}
\rho_{\textcolor{black}{swarm}}(t_0) = U S V^{\dag},\,\,\rho_{\textcolor{black}{swarm}}(t_1) = U' S' V'^{\dag},
\end{equation}
\textcolor{black}{where $S$ (respectively $S'$) is a diagonal matrix with the singular values of $\rho_{\textcolor{black}{swarm}}(t_0)$ (respectively $\rho_{\textcolor{black}{swarm}}(t_1)$).
These new matrices can be used to reconstruct the process from $\rho_{\textcolor{black}{swarm}}(t_0)$ to $\rho_{\textcolor{black}{swarm}}(t_1)$. In fact, the needed time-evolution operator is:}
\begin{equation}
O = U'(V'^{\dag})^{-1}VU^{\dag}.
\end{equation}
The choice of the dynamics is fundamental to shape the global motion of a swarm. \ma{Here,} the robots exchange information about individual target proximity, and then the overall swarm reaches a point in space, \ma{e.g., toward a search and rescue scenario where the robot swarm needs to identify and reach targets for performing rescue operations}. The point in space is represented as an ideal swarm centered on it. In our case, it can be the target density matrix. The approximation of the time operator represented the required overall transformations to let the swarm approximately reach that region.
We can also apply to $\rho_{\textcolor{black}{swarm}}(t_0)$ some given specific dynamics, and observe how the swarm is moving accordingly. A classic example is $U = e^{- i H t}$.
The systematic definition of density matrices for a swarm, and a set of subsequent computations of the dynamics between pairs of time points, can allow one to approximate the functions inside the dynamics. Such an approach can foster precise applications in the domain of global-to-local control of the swarms.

\begin{figure}
    \centering
    \includegraphics[width=\textwidth]{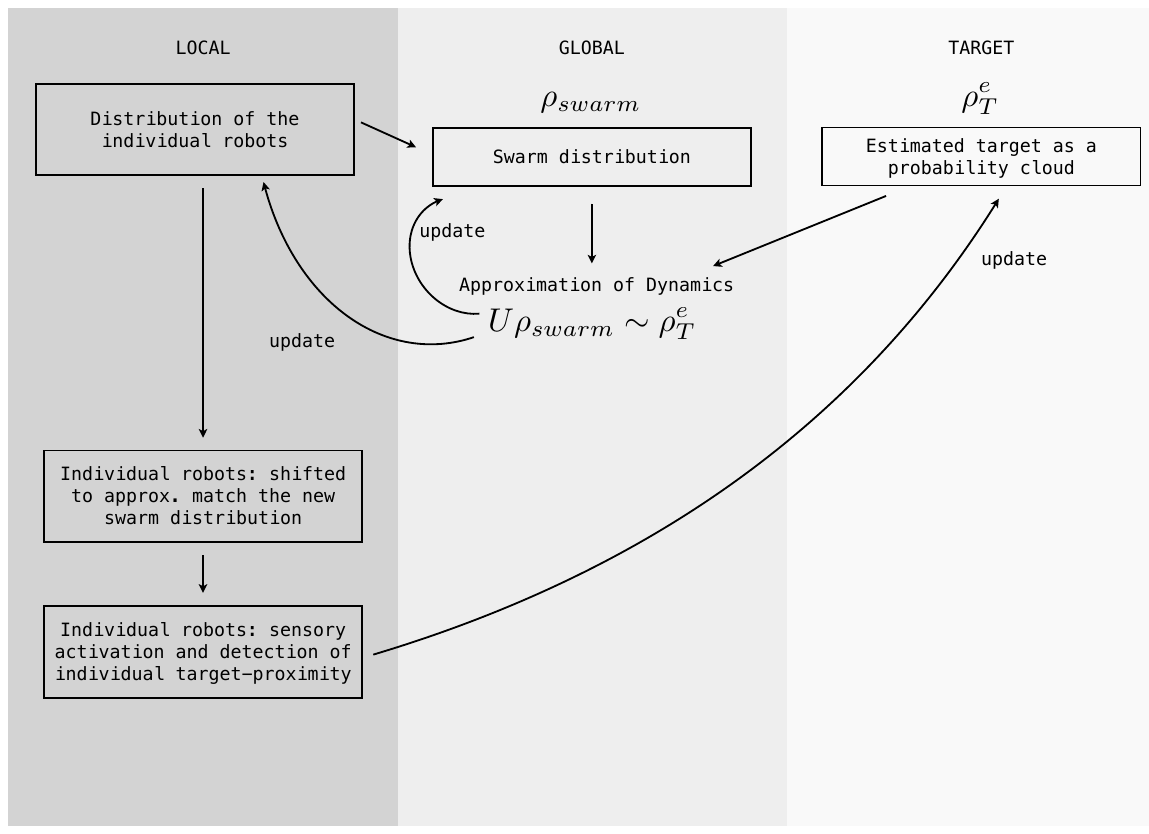}
    \caption{An algorithm for swarm-robotic target reaching based on the defined matrices, and including an interaction between local and global.}
    \label{algo}
\end{figure}

In particular, we can define the algorithm illustrated in Figure \ref{algo} \textcolor{black}{and in Algorithm \ref{algo2}}. Our formalism can thus give some theoretical prediction on target reaching that can be experimentally verified.
Let us describe the idea of Figure \ref{algo}. We start with the local dimension, defining an initial distribution of robots in terms of $|\psi_i\rangle$. Then we compute their density matrix distribution, thus going into the global dimension. We also need a target position defined as an estimated probability density, \ma{denoted} $\rho_T^e$. This information will be progressively  refined. At this point \textcolor{black}{we can} approximate a dynamics $U$ such as $U\rho_{swarm}\sim\rho_T^e$. We can update the swarm distribution in terms of density matrix. And we also have to update the distribution of robots at the local dimension. Since more individual configuration correspond to the same density matrix distribution, the robots can be shifted with the smallest displacements such that their $\psi_i$ approximate the new $\rho_{swarm}$. The robots activate their sensors, and detect their individual proximity to the target. This information allow us to update the target distribution $\rho_T^e$, which allows for a new dynamics computation. The cycle continues until the swarm reaches the target with a great precision. This method, inspired by optimization techniques, can help yield prediction on the number of passages for target reaching. \textcolor{black}{In the algorithm, we adopt the notation $\rho_i$, $\rho_f$ to denote the density matrix at the beginning of arena exploration and at the end of it, respectively. We do not use here $t_0$ and $t_1$ because we prefer to not make hypothesis here on the number of time points required to reach the target.}

\textcolor{black}{
\begin{algorithm}[ht!]
\caption{Distributed swarm algorithm.}\label{algo2}
\begin{algorithmic}[1]
\Require $N$ robots with states $\{\psi_i\}_{i=1}^N$, expected target density $\rho_T$, arena target position $T$
\State \textbf{Initialize:} robots' positions, individual quantum states
\Repeat
    \State \textbf{(A) Local $\rightarrow$ Global:} each robot $i$ sends its local classical position $p_i$ and its quantum state $\rho_i = |\psi_i\rangle\langle\psi_i|$ to the global node
    \State \textbf{(B) Global aggregation:} compute swarm density matrix $\rho_{swarm}$
    \State \textbf{(B) Global dynamics estimation:} use $\rho_{\text{swarm}}$ and the estimated target cloud to compute an approximate dynamics $U$ driving $\rho_{\text{swarm}}\rightarrow\rho_T$
    \State \textbf{(B) Send dynamics to individuals:} determine per-robot quantum updates and transmit them
    \State \textbf{(C) Classical motion plan (global $\rightarrow$ local):} computation of the global desired swarm spatial distribution approximating the estimated target location; instructions sent to the individuals
    \For{each robot $i$ (local)}
        \State Robot $i$ activates sensors to estimate target-proximity
        \State Robot $i$: classical motion, quantum-update received:
        \[
            \psi_i \leftarrow \text{normalize}(\psi_i + \gamma_i(\psi_T - \psi_i))
        \]
        $\gamma_i$ is computed globally
    \EndFor
    \State \textbf{(A) Repeat:} 
\Until{trace\_distance$(\rho_{\text{swarm}}, \rho_T)$ below tolerance \textbf{or} max iterations reached}
\end{algorithmic}
\end{algorithm}
}

\clearpage
\newpage

\subsection{\ma{Distributed Quantum Swarm}}

\textcolor{black}{We end this Section with an example of possible implementation of Algorithm \ref{algo2}, without an explicit approximation of the time-evolution operator, but including the effects of the dynamics on the robotic movements. We consider the same swarm and scenario as seen in Subsection \ref{example_qiskit}.
We implement here the A/B/C loop, where A indicates the dimension of local robots, B the overall swarm, and C the target, and each of them is described by density matrices (or approximate density matrices, as for the target). For the sake of simplicity, we consider here an omniscient scenario, as in \cite{swarm_paper}, and an exchange between local and global dimensions of the swarm \cite{maria_transactions}.
The implementation contains the following steps:
\begin{itemize}
\item robots send positions and local $\rho$ to the global node;
\item global dimension: computation of aggregates $\rho_{swarm}$ (with per-robot mixing factors) and approximation of the dynamics;
\item global dimension: sends these mixing factors and a desired spatial distribution (target cloud) to the individual robots;
\item robots move according to their local sensors to match the desired distribution and apply quantum updates;
\item Finally, as output, each robot returns the states of the final robots and the final $\rho_{swarm}$.
\end{itemize}
Here, the `approximate dynamics' $U$ is implemented as a small convex mixing of each local $\psi$ toward $\psi_T$ of the target, with individual mixing coefficient obtained from fidelity $F(\rho, \sigma) = \left(tr\sqrt{\sqrt{\rho}\sigma\sqrt{\rho}}\right)^2$, where $\rho,\,\sigma$ are the density matrices of two quantum states, but it can also be done with trace distance.
We implement the discussed time evolution in three different ways:
\begin{itemize}
\item via a \textit{Classical–Quantum Hybrid Iteration}, constituted by a dissipative quantum relaxation \cite{lindblad} and a consensus-type movement rule \cite{olfati} for the robots. Each robot of the swarm has a local, pure quantum state $\psi_i$ and moves classically toward the target through the space. The global $\rho_{\text{swarm}}$ is updated iteratively, according to local information and a small, quantum-inspired feedback term (a Lindblad-like dissipative operator \cite{lindblad}), according to the update rule:
\begin{align}
\rho_{\text{swarm}}(t+1)
&= \rho_{\text{swarm}}(t) + \alpha \big( \rho_T - \rho_{\text{swarm}}(t) \big)
   + \mathcal{L}_{\text{noise}}\!\big[\rho_{\text{swarm}}(t)\big], \\[4pt]
(x_i, y_i)_{t+1}
&= (x_i, y_i)_t + \beta \sum_j w_{ij}\,\big((x_j, y_j)_t - (x_i, y_i)_t\big);
\end{align}
\item via a \textit{Completely Positive Trace-Preserving Map (CPTP)} Update \cite{nielsen, lindblad2}, a quantum map that replaces the additive update, according to the update rule: 
\begin{align}
\rho_{\text{swarm}}(t+1)
&= (1 - \eta)\,\rho_{\text{swarm}}(t) + \eta\,\rho_T, \\[4pt]
(x_i, y_i)_{t+1}
&= (1 - \beta)\,(x_i, y_i)_t + \beta\,(x_T, y_T),
\end{align}
\item finally, with the \textit{Pure CPTP Global Update} \cite{ruskai, lindblad2, carlen}, with a direct projection toward the target state. For a specific choice of parameters, one obtains an instantaneous projection (and a trivial convergence). The update rule is:
\begin{align}
\rho_{\text{swarm}}(t+1)
&= (1 - \eta)\,\rho_{\text{swarm}}(t) + \eta\,\rho_T, \\[4pt]
(x_i, y_i)_{t+1}
&= (1 - \beta)\,(x_i, y_i)_t + \beta\,(x_T, y_T),
\end{align}
and, in the case of for $\eta = \beta = 1$, we obtain:
\[
\rho_{\text{swarm}}(t+1) = \rho_T, \qquad
(x_i, y_i)_{t+1} = (x_T, y_T),
\]
\end{itemize}
Finally, we define control parameters such as the number of iterations and the tolerance.}

\textcolor{black}{Let us start with the first method. 
And the robotic sensors are here simplified as a measure of individual proximity to the target.
The mixing coefficient is larger when fidelity is low or the global distance is high. For the overall spatial distribution, we consider a Gaussian cloud centered at target, with a $\sigma$ depending on the global distance.}
\textcolor{black}{At this point, the updates are sent to the individuals. They concern the mixing factors and the cloud parameters. The individual robots receive them and then move on the arena accordingly.}
\textcolor{black}{The overall procedure is repeated until the target is reached, or when a maximum number of iteration is attained.}

\textcolor{black}{For our implementation, we consider the previously-defined ten robots. The initial trace distance of $0.74$ decreases to $0.24$ after 20 iterations, $0.18$ after 30 iterations, and, after only 5 iterations, of $0.54$.
Apparently, this method may appear slow. However, one should take into account that no quantum-implemented logic gate has been used, avoiding the time of its implementation on a quantum circuit, as it was done in \cite{swarm_paper}, and there was no need for a signal transmission between the individual robots and a computational central.}

\textcolor{black}{We can then implement the two other techniques.
As aforementioned, the convergence of the second one, the pure CPTP, is trivial: it only corresponds to equating the density matrix of the swarm to the one of the target. For smaller values of $\alpha$, it is visible a progressive approach of the swarm. With the initial distance of $0.74019942$, after one iteration we get $0.44$, and after two, we get $0.05$. We show the results of the comparison between the three techniques in Table \ref{comparisons_convergence_speed}.
}

\begin{table}[ht!]
\caption{Comparison of the convergence speed as obtained distance between $\rho_{swarm}$ and $\rho_T$ for some variants of Algorithm \ref{algo2}, for the initial distance $\rho_{swarm},\,\rho_T = 0.740199$}\label{comparisons_convergence_speed}
\begin{tabular}{c|ccc}
\hline
 & 5 iterations & 20 iterations & 30 iterations \\
\hline
quantum-classical hybrid & 0.54 & 0.24 & 0.18 \\
CPTP only & 0.47 & 0.40 & 0.40 \\
Pure CPTP Global Update & 0.0072 & 0.0000031 & 0.00000002 \\
\hline
\end{tabular}
\end{table}

\noindent \textcolor{black}{
\textbf{Quantum Swarm in the Real World. } In a concrete scenario, nodes move inside a defined 2D Geographical Area ($GA$), whose dimensions are $X$ and $Y$. At the application layer, we can imagine that robots can exchange few coordination packets:
\begin{itemize}
\item Status and health messages: generally used to maintain awarness of each robot condition (status update, error alert, low battery, etc.);
\item Localization and mapping messages: generally used for exploration (position update, map fragment, landmark detection, etc.);
\item Task coordination messages: used to distribute and synchronize tasks among robots (task announce, task claim, task status, task complete, etc.);
\item Motion management and collision avoidance: for ensuring safe movements and swarm keeping (move willing, obstacle detection, formation update, etc.);
\item General information sharing: used for distributed consensus or decision-making;
\item roles management: used to assign specialized roles to different nodes (resource available, role assignment, load transfer);
\item Global coordination: generally used for mission flow control (mission start, mission pause, mission complete, global alert).
\end{itemize}
The description above is referred to just one of the possible scenario. Our proposed approach is completely general, so the set of provided messages can be stretched or extended.
}

\textcolor{black}{We believe that the physical layer can be contextualized in function of the considered environment. In particular, RF waves can be used in free space, acoustic waves can be considered for underwater missions, etc. In particular, a swarm primarily utilizes wireless communication for inter-robot data transmission. Then, depending on mission requirements and deployment environments, one or more of the following technologies may be employed:
\begin{itemize}
\item Wi-Fi (IEEE 802.11 protocols  family):
it provides high data rates suitable for rich information exchange such as maps, images, or sensor data. Common in indoor or short-range outdoor environments;
\item ZigBee (IEEE 802.15.4 protocols family): generally used for low-power, low-bandwidth communication where energy efficiency and robustness are prioritized over data throughput. It is suitable for large-scale swarms with lightweight message exchanges;
\item Bluetooth: it is more appropriate for small teams or close-proximity coordination with minimal power consumption;
\item LoRa or Sub-GHz Links (optional in long-range missions): these technologies are used when the swarm must cover large outdoor areas with sparse connectivity and low data rate requirements.
\end{itemize}
}

\section{Conclusions and Discussions}\label{future}
In this article, we proposed  a new modeling methodology for a quantum robotic swarm using density matrices to describe swarms of micro and nano robots under quantum effects in a compact yet precise manner. We also proposed some computational examples illustrating applications of our modeling formalism. \ma{The proposed framework is illustrated in both small and large robotic swarms, ranging from swarms of $2$ robots upto $1000$ robots. As the number of the robots increase, the shape of the cloud at each time instant is different. However, we show that the size of the obtained density matrices remain constant, dependent only on the degrees of freedom of the robots. This presents a significant computational advantage over previous approaches such as in~\cite{swarm_paper}. In addition, we have proposed algorithms and implementations for target-reaching, and analyzed the obtained results. For toy robotic swarms with $2$ or $3$ robots, the computations are straightforward and thus performed manually. For more complex computations, such as larger swarms and density matrix computations, the open-source software Qiskit\footnote{\url{https://www.ibm.com/quantum/qiskit}} is used. Finally, a brief discussion on telecommunication tools and features, as a further step toward a more realistic and real-life implementation of our model is presented. }


\bl{In the future, one may consider the application of our approach to describe the motion of single robots as well as the center of mass of the quantum swarm. Experiments on physical robots may also help investigate the limits of our approach,} especially concerning the necessary calibration of parameters representing divergence and convergence, which also depend upon the size of the swarm, the amplitude of movements of their components. Future research can focus on the matrix definition of multiple swarms and hierarchical approaches. In fact, we can define the density matrix of a system with multiple swarms as the tensor product of the density matrices of each swarm, or as the density matrix of the whole swarm seen as a mixed state of individual swarms. The information on the single swarm can vice versa be recovered performing the operation of partial trace with respect to all the other swarms, considered as ``environment,'' to the overall density matrix. 

\textcolor{black}{Further studies can exploit the density-matrix based formalism to develop control approaches for quantum robots \cite{emzir}.}
\textcolor{black}{In fact, our formalism with density matrices for robotics swarms can pave the way toward future applications of stability criteria to the robotics swarms, adapting control criteria defined for open quantum systems to the robots} \cite{emzir}. Moreover, one can also address the matrix depiction of multi-layer systems, recently used to characterize hierarchical systems of swarms communicating with mobile computing centrals \cite{wivace}, whose information-sharing systems are shaped according to the protocols of telecommunications \cite{VTC}.

\section*{Funding}

The work by M. A. is funded by the German Research Foundation (DFG, Deutsche Forschungsgemeinschaft) as part of Germany's Excellence Strategy – EXC 2050/1 – Project ID 390696704 – Cluster of Excellence ``Centre for Tactile Internet with Human-in-the-Loop'' (CeTI) of Technische Universität Dresden.

\section*{Code availability}

\textcolor{black}{The codes in Python to implement the proposed examples can be accessed at \url{https://github.com/medusamedusa/density_matrix_for_swarms}.}

\bibliographystyle{plain}
\bibliography{egbib}

\end{document}